\documentclass[11pt]{article}
\usepackage[margin=1in]{geometry}
\usepackage{graphicx,booktabs,amsmath,amssymb,microtype,array,float}
\usepackage{xurl}          %
\usepackage[hidelinks]{hyperref}
\usepackage[numbers,sort&compress]{natbib}
\usepackage{caption}
\title{Do Reviewers Still Reward Lexical Complexity?\\
A Frozen-Rater Study of Preference Drift in 124K ICLR Reviews}
\author{Jiabin Zheng\\[2pt]
{\small School of Computer Science, Peking University \quad \texttt{jiabinzheng@pku.edu.cn}}}
\date{}

\begin{document}
\maketitle

\begin{abstract}
\noindent
Large language models have collapsed the cost of producing lexically elaborate prose, and
whether peer reviewers still reward it is a question about the evaluator, not about the
text. When the association between a writing cue and review scores moves across years, the
reviewers may have changed, the submissions may have changed, or both, and a regression of
scores on text cannot say which. We separate the two with a frozen rater: $81{,}850$ machine
reviews of ICLR submissions from 2018 to 2025, all generated in one February--April 2025
window with one model family and one prompt, so that its year-to-year coefficients track
submission composition alone and the human-minus-frozen trend difference identifies
reviewer preference drift. On $32{,}638$ submissions with $124{,}615$ human reviews, the
human coefficient on non-domain lexical complexity falls from $+0.142$ to $-0.015$ while the
frozen rater moves from $+0.080$ to $+0.082$; the three-way difference-in-differences is
$-0.0100$ ($q=0.013$), and forty random-wordlist placebos through the same specification
centre on zero. Humans still reward sentence-length variability, which the frozen rater
never registers, while the frozen rater still pays for lexical complexity at its earlier
rate. Every claim is held to a double gate of false-discovery control and interval
exclusion, and the findings that failed adversarial re-testing are reported. Reviewers
discounted a cue whose production cost collapsed, as models of manipulable signals
prescribe; an LLM judge calibrated to historical human preferences inherits the earlier
schedule and drifts out of alignment while its agreement with humans on totals stays
ordinary.
\end{abstract}

\section{Introduction}
\label{sec:intro}

Large language models now write and rewrite a measurable share of scientific text. The
share of ICLR reviews with substantial machine-modified content rose sharply after the
release of ChatGPT \citep{liang2024monitoring}, the vocabulary of biomedical abstracts
shifted toward a recognisable set of style words within a year \citep{kobak2025delving},
and the use of models in computer-science manuscripts grew faster than in any other field
\citep{liang2024mapping,kousha2026language}. The evaluation side of the same system is
less documented. Reviewers form scores partly from what a paper looks like on the page, and
a literature has estimated how much each presentation cue is worth
\citep{armstrong1980unintelligible,oppenheimer2006,lu2019impact,martinez2021jargon}. What is
not known is whether human reviewers changed what they reward once one of those cues,
lexically elaborate prose, stopped costing an author's time and started costing a prompt.

\begin{figure}[t!]
\centering
\includegraphics[width=\textwidth]{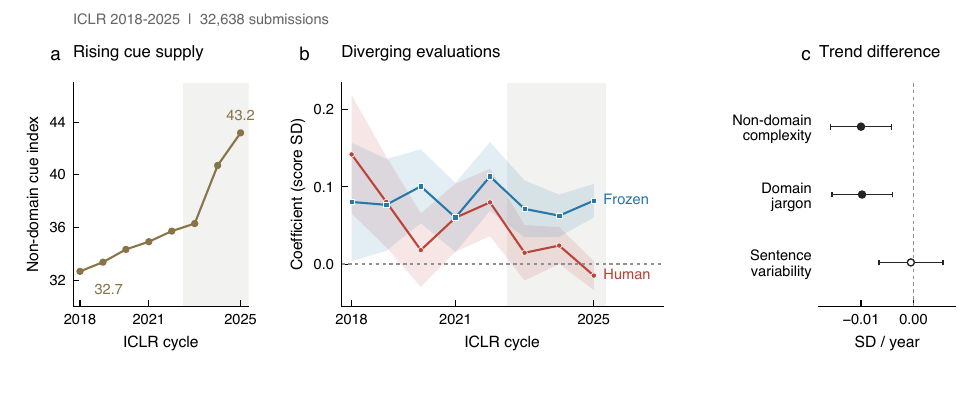}
\caption{\textbf{Rising cue supply, diverging evaluations.} (a) Mean non-domain lexical
complexity in submitted abstracts. (b) Human and frozen-rater cue-score coefficients by ICLR
cycle, with 95\% confidence bands. (c) Annual human-minus-frozen trend differences for three
cues, with 90\% intervals. Filled markers pass both BH-FDR $q<0.10$ and interval exclusion;
open markers do not. Shading denotes the 2023--2025 cycles, not an estimated treatment effect.}
\label{fig:main}
\end{figure}

The question matters beyond peer review. If a cue was informative because it was costly
\citep{spence1973signaling}, its value as a signal should erode once a cheap substitute
exists; models of evaluation under manipulable signals predict that receivers underweight
such cues, and underweight them more as manipulation spreads
\citep{frankel2019muddled,ball2025scoring,hennessy2023goodhart}. Whether reviewers actually
responded is an empirical fact about human evaluators adapting to a cost shock, and it is
also a calibration fact for any automated judge trained or prompted to mimic historical
human preferences \citep{zheng2023judging,ye2025justice}. The difficulty is that the natural
design cannot answer it. Regressing scores on a text feature year by year and reading the
trend confounds three movements: the evaluator's weight on the cue may have changed; the
distribution of submitted text has certainly changed; and the papers that carry the cue may
have changed in other respects. A falling coefficient is consistent with all three, and the
text alone cannot say which \citep{jung2025acceptance}.

\begin{figure}[t]
\centering
\includegraphics[width=\textwidth]{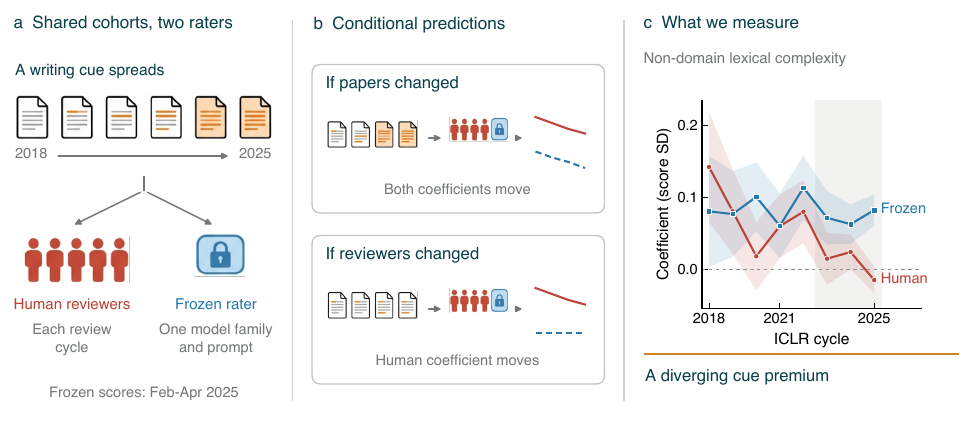}
\caption{\textbf{A fixed comparator for changing evaluation.} (a) Submission cohorts receive
contemporaneous human reviews and retrospective scores from one model family and prompt.
(b) Schematic predictions for shared composition effects versus human-specific change,
conditional on comparable composition responses across raters. (c) Observed non-domain lexical
complexity coefficients on a common scale, with 95\% confidence bands; the two arms use their
available ratings, with different annual sample sizes. Shading denotes the 2023--2025 cycles.
Panels a--b are conceptual. A fixed rater alone does not establish the comparability condition
needed to interpret divergence as preference drift.}
\label{fig:design}
\end{figure}

We resolve the confound with a rater whose standard is known not to have moved. The
Gen-Review corpus \citep{demetrio2025genreview} contains $81{,}850$ machine-generated
reviews of ICLR submissions from 2018 to 2025, all produced in a single February--April 2025
window, from one model family, under one prompt (Figure~\ref{fig:design}). That rater read the 2018 papers and the
2025 papers with the same instrument, so its year-to-year coefficient movement carries
composition change and nothing else, and the human trend minus its trend identifies the part
that belongs to the humans. This is the anchor-item logic of non-uniform differential item
functioning \citep{swaminathan1990detecting,kopf2015anchor}, implemented as a three-way
difference-in-differences (feature $\times$ year $\times$ human) on the stacked human and
machine observations of the same papers. The machine is used as a fixed ruler, never as
ground truth: it may be badly calibrated, provided it is calibrated the same way every year.

We apply the design to $32{,}638$ submissions and $124{,}615$ human reviews. Four text cues
are measured on the abstract as it was submitted rather than as it stands today, a
distinction that turned out to matter (Section~\ref{sec:data}); lexical complexity is split
into the general scientific vocabulary that carries style and the domain jargon that
carries topic, because an undifferentiated list measures topic; and every claim is held to
a double gate of false-discovery control and interval exclusion, with forty random-wordlist
placebos run through the identical specification. The human weight on non-domain lexical
complexity falls to zero while the frozen rater's does not move (Figure~\ref{fig:main}); in
the late window the frozen rater still pays for the cue at its earlier rate and never
registers the sentence-length variability that humans still reward; the placebos centre on
zero; and seven findings that did not survive adversarial re-testing are reported rather
than dropped.

Three questions organise the study: whether human reviewers' weight on lexical complexity
changed between 2018 and 2025 once composition change is removed by the frozen rater
(RQ1); whether the change is specific to the cue whose cost collapsed or a general
shrinkage of every text coefficient (RQ2); and what a rater frozen at a pre-2023 standard
still rewards in 2023--2025 that human reviewers no longer do (RQ3).

The paper makes three contributions.
\begin{itemize}
\item \textbf{An identification strategy.} The frozen-rater design separates preference
drift from composition drift wherever a fixed-prompt rater has been applied
retrospectively to a corpus; it requires only that the rater not have been re-calibrated,
not that it be right.
\item \textbf{The measurement.} The human coefficient on non-domain lexical complexity
falls from $+0.142$ to $-0.015$ while the frozen rater holds at $+0.080 \to +0.082$, a
divergence that survives forty random-wordlist placebos, a year permutation and a
pre-period window, and that is absent for the sentence-length variability humans still
reward.
\item \textbf{A data-provenance correction.} The released abstracts were post-revision
text; revision is more common for accepted papers and in later years; rebuilding the corpus
on submitted versions made the decline steeper rather than weaker.
\end{itemize}

Substantively, reviewers discounted a cue whose production cost collapsed, while the frozen
rater kept paying for it. That the association moved has been reported as a correlation on
this venue \citep{jung2025acceptance}; what was missing was a way to say whose movement it
was. The design therefore shifts the question from whether reviewers reward lexical
complexity to whose standard moved, and it carries a warning for any LLM judge calibrated to
a historical human preference: it drifts out of alignment silently while its agreement with
humans on totals stays ordinary.

\section{Related work}
\label{sec:related}

\paragraph{Observed shifts, and what remained unidentified.}
A process-centric study of ICLR 2017--2025 reports that ``the notion of good writing has
shifted'': readability indices correlate with acceptance one way before 2024 and another
way after \citep{jung2025acceptance}. That is an observation of the trajectory we study, and
it is unidentified in the sense of Section~\ref{sec:ident}: two period bins, univariate
correlations with acceptance, and no fixed comparator. We replicated its design on our
abstracts (released with the analysis). In the pre-period the sign of the readability
correlation is the reverse of the full-text report; in the post-period every readability
correlation is indistinguishable from zero; and once team size and the other indices are
controlled, no readability index retains significance while log author count carries a
coefficient an order of magnitude larger. We take the shift as observed and the
identification as our contribution. In the scientometric literature, linguistic complexity
has been related to impact and to disciplinary writing style
\citep{lu2019writing,lu2019impact}, jargon to citation loss \citep{martinez2021jargon},
and the readability of abstracts to a century-long decline driven by general scientific
vocabulary \citep{plavensigray2017}; none of these designs separates the evaluator from the
text.

\paragraph{Machine raters as instruments, not as judges.}
A large literature asks whether an LLM reproduces or improves on human review, and
documents the biases of LLM judges: verbosity and position effects
\citep{zheng2023judging,dubois2024length}, unfairness across candidates
\citep{wang2024fair}, cognitive biases \citep{koo2024cognitive}, self-preference
\citep{panickssery2024self}, and a catalogue of hidden biases in LLM-assisted review
\citep{ye2025justice}; a survey of LLM-based review notes that such raters ``may reward
clear structure and polished prose while overemphasizing stylistic signals relative to
substantive contribution'' \citep{nguyen2026survey}, which is the frozen half of our result
stated as a general concern, and controlled edits show that LLM reviewers are manipulable by
surface changes \citep{ye2024risks} while models asked only to correct grammar still alter
meaning \citep{abdulhai2026distort}. Our use is orthogonal and deliberately weaker: we
require only that the machine rater's \emph{own} behaviour be constant across the years it
scores, not that it be right. Two recent preprints attack the mirror-image problem, whether
a change in a monitored score comes from the system or from a drifting LLM judge, and
anchor the judge to frozen human labels \citep{li2026drifted,yang2026judge}; we anchor the
humans to a frozen judge. Large-scale field experiments have established that the human
reviewer pool itself is being reshaped by model use
\citep{liang2024monitoring,russo2025lottery,thakkar2026randomized}, and the ideation study
of \citet{si2025ideas} is the closest human-rater benchmark in scale.

\paragraph{Peer-review experiments and measurement invariance.}
Controlled studies of peer review have identified reviewer biases by design rather than by
correlation \citep{tomkins2017reviewer,stelmakh2019testing,goldberg2025peer,shah2022challenges},
and readability standards have been shown to be applied unevenly in review
\citep{hengel2022publishing}. Psychometrics supplies the vocabulary for our estimand: when
the relationship between a latent trait and an indicator differs across groups that is
differential item functioning; when the difference is in the slope it is non-uniform DIF;
and detecting it requires anchor items assumed invariant
\citep{mellenbergh1989item,holland1993dif,swaminathan1990detecting,kopf2015anchor}. We adopt
that vocabulary rather than coining terms.

\paragraph{Signal debasement.}
Economics supplies the interpretation. A cue is informative because it is costly
\citep{spence1973signaling}; when a sender can manipulate the feature a receiver observes,
the receiver's optimal rule underweights it relative to its raw predictive content, and
underweights it more as manipulation becomes cheaper or more heterogeneous across senders.
\citet{frankel2019muddled} call the resulting equilibrium object muddled information;
\citet{ball2025scoring} shows the optimal score deliberately underweights manipulable
features; \citet{hennessy2023goodhart} describe the comparative static as slopes shifting
downward, the general form of Goodhart's law \citep{goodhart1984problems}. The same logic
appears in machine learning as strategic classification and performative prediction
\citep{hardt2016strategic,perdomo2020performative}. The prediction is directional and it is
about slopes, which is what we measure. We present it \emph{after} the design and the
estimates, so that debasement is an interpretation of an identified shift rather than an
assumption imported to produce one.

\section{Method}
\label{sec:method}

\subsection{Panel and outcome}
\label{sec:data}

\paragraph{The panel covers eight ICLR cycles scored by both arms.}
Table~\ref{tab:corpus} describes the corpus: $32{,}638$ ICLR submissions from 2018 to
2025, with $124{,}615$ human reviews and $81{,}850$ frozen machine reviews. Per-year analysis
samples run from $922$ papers (2018) to $11{,}517$ (2025) in the human arm and from $925$ to
$8{,}375$ in the machine arm; the gap is machine-review coverage, not a different paper set.
Human reviews and decisions come from the venue's public record; machine reviews are the
released Gen-Review corpus \citep{demetrio2025genreview}.

\begin{table}[t]
\centering
\scriptsize
\setlength{\tabcolsep}{4pt}
\begin{tabular}{lrrrrrrr}
\toprule
Year & Submissions & Human arm & Frozen arm & Non-domain & Domain jargon & Sent.-length SD & Abstract words \\
\midrule
2018 & $935$ & $922$ & $925$ & 32.7 & 47.4 & 7.47 & 161 \\
2019 & $1{,}419$ & $1{,}419$ & $1{,}390$ & 33.4 & 48.7 & 7.78 & 161 \\
2020 & $2{,}213$ & $2{,}213$ & $2{,}175$ & 34.3 & 48.5 & 7.58 & 163 \\
2021 & $2{,}594$ & $2{,}594$ & $2{,}526$ & 34.9 & 49.3 & 7.68 & 171 \\
2022 & $2{,}618$ & $2{,}618$ & $2{,}522$ & 35.7 & 48.6 & 7.72 & 178 \\
2023 & $3{,}797$ & $3{,}797$ & $3{,}662$ & 36.3 & 48.7 & 7.85 & 181 \\
2024 & $7{,}401$ & $7{,}262$ & $5{,}607$ & 40.7 & 49.9 & 7.71 & 186 \\
2025 & $11{,}667$ & $11{,}517$ & $8{,}375$ & 43.2 & 51.3 & 7.60 & 189 \\
\bottomrule
\end{tabular}

\caption{\textbf{Corpus by submission year.} Submissions with text features; papers with
human ratings; papers with frozen machine ratings; and the corpus mean of each cue in the
submitted abstract. Non-domain lexical complexity and domain jargon are syllable mass per
word of the respective word class; sentence-length SD is in words.}
\label{tab:corpus}
\end{table}

\paragraph{The outcome is the within-year standardised rating.}
The outcome is the reviewer rating, standardised within year. We standardise within year
because the estimand is \emph{relative} cue use, and because a year-level shift in the
scoring scale is a documented failure mode of longitudinal rating comparisons. We cluster
standard errors by paper, since a paper contributes several reviews
\citep{cameron2015practitioner}.

\subsection{Submitted-version abstracts}
\label{sec:versions}

\paragraph{The released abstracts are not the abstracts reviewers read.}
The released corpus carries each abstract as it stands on the venue's site today, and that
is not the abstract reviewers read: submissions are revised during discussion and, if
accepted, again for camera-ready. We audited this against the venue's version histories on
a stratified sample of $1{,}800$ submissions from 2018--2023. Among abstracts that changed
between first and last version, the corpus text matched the last version in every case and
the first in none; the share that changed rose from $12\%$ of submissions in 2018 to
$72\%$ in 2021, was higher for accepted papers, and the edits were substantive (median
similarity to the submitted text $0.73$). Revision is more common for accepted papers and
in later years, so this contamination is correlated with both the outcome and the trend.
We therefore rebuilt the abstract layer on submitted versions (Table~\ref{tab:versions} in
Appendix~\ref{app:versions}): the first archived version for 2018--2023, a public
pre-review scrape for 2025, and for 2024, where the venue's current API no longer exposes
the original, the earliest public snapshot, taken after discussion but before camera-ready.
Every number in this paper is computed on the rebuilt layer. The correction moves each
late-year human coefficient further from zero and leaves the frozen arm unchanged: the
early-to-late human contrast on non-domain lexical complexity is $-0.0433$ on current text
and $-0.0485$ on submitted text. The contamination had been biasing the headline toward the
null. The machine arm read the PDFs as the corpus authors downloaded them, that is, current
versions; we return to this in Section~\ref{sec:discussion}.

\paragraph{Abstracts, because every full-text route is selected on the outcome.}
Our cues are computed on submitted-version abstracts while the machine rater read full PDFs.
This is a real limitation and we chose it over a worse alternative: we measured four routes
to conference full text and every one is either outcome-correlated or unusable. Matching to
arXiv skews the sample $+25$ percentage points toward accepted papers; OpenAlex skews it
$+42$; the venue's own API is unbiased but rate-limited to $36$ PDFs per hour, which is
$38$ days for this corpus. We report the abstract-only result rather than a full-text result
computed on a sample selected on the outcome.

\subsection{Cues}
\label{sec:cues}

\paragraph{Four cues, mutually controlled.}
Table~\ref{tab:features} defines the four cues. Lexical complexity is
measured as the syllable mass of words with three or more syllables, the complex-word count
of the Fog index \citep{gunning1952} and the manipulation of \citet{oppenheimer2006}, and
it is split by whether the word belongs to a domain lexicon. Words outside the lexicon are
the general scientific vocabulary whose growth drives the long-run decline in the
readability of abstracts \citep{plavensigray2017} and whose recent surge marks
machine-assisted writing \citep{kobak2025delving}; we call this cue non-domain lexical
complexity. Words inside the lexicon are domain jargon, the specialised terminology that
carries topic and has its own citation consequences \citep{martinez2021jargon}. The split
is not cosmetic. It is a direct consequence of an error we made and had to withdraw
(Section~\ref{sec:retractions}): an undifferentiated complexity or hype list is dominated by
ordinary machine-learning vocabulary and measures topic, not style. Sentence-length SD is the
syntactic half of linguistic complexity in the sense of \citet{lu2019writing}; abstract
length is a control, since length is the best-known confound of automated judgement
\citep{dubois2024length}. Both wordlists and their hit distributions are released.

\begin{table}[t]
\centering
\small
\setlength{\tabcolsep}{4pt}
\begin{tabular}{p{3.0cm}p{4.1cm}p{5.0cm}p{2.6cm}}
\toprule
Cue (this paper) & Construct in the literature & Operationalisation on the submitted abstract & Sources \\
\midrule
Non-domain lexical complexity & general scientific vocabulary; the ``complex words'' of readability formulas outside the field's own terminology & syllable mass of words with three or more syllables that are absent from the domain lexicon, per word of text & \citet{gunning1952}; \citet{oppenheimer2006}; \citet{plavensigray2017} \\
Domain jargon & domain-specific terminology & the same count for words present in the domain lexicon & \citet{plavensigray2017}; \citet{martinez2021jargon} \\
Sentence-length SD & syntactic variability across sentences & standard deviation of sentence length in words & \citet{lu2019writing} \\
Abstract length & control & number of words & --- \\
\bottomrule
\end{tabular}

\caption{\textbf{Cues and their operationalisation.} Each cue is computed on the abstract as
submitted, standardised within year, and entered jointly with the others.}
\label{tab:features}
\end{table}

We take syllable counts from the CMU pronouncing dictionary rather than from hyphenation,
which splits words at typographic break points and undercounts syllables systematically, and
we calibrated the implementation against a published reference passage, obtaining $37.46$
against a reference value of $37.5$. Readability values from hyphenation-based libraries are
therefore not comparable to ours.

\begin{figure}[t]
\centering
\includegraphics[width=0.88\textwidth]{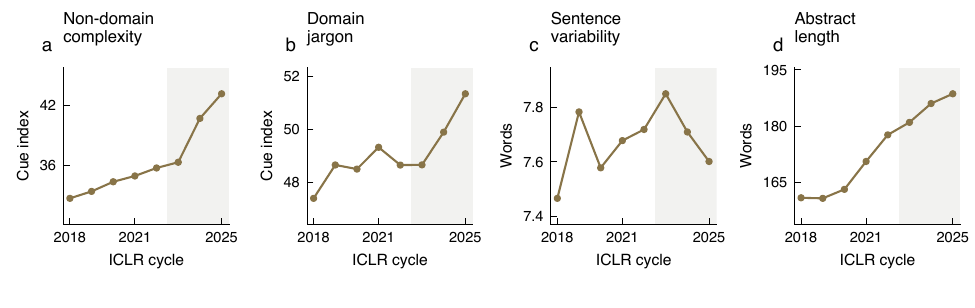}
\caption{\textbf{Composition drift in the abstract layer.} Corpus means by ICLR cycle. The
first two panels show the lexical-complexity indices; sentence-length variability and abstract
length are measured in words. Shading marks the 2023--2025 cycles.}
\label{fig:drift}
\end{figure}

\paragraph{Composition drift is real.}
Figure~\ref{fig:drift} and Table~\ref{tab:corpus} show why the identification problem is
not academic. The corpus mean of non-domain lexical complexity rises from $32.67$ to
$43.18$, sharply after 2023; domain jargon rises from $47.39$ to $51.34$; abstract length
rises from $161$ to $189$ words. Sentence-length variability moves the other way after
2023, from $7.85$ down to $7.60$. The cue humans still reward is becoming scarcer in the
corpus while the cue they stopped rewarding is becoming abundant, which is the pattern
independent corpus studies report for machine-assisted writing
\citep{kobak2025delving,kousha2026language}.

\subsection{Identification}
\label{sec:ident}

\paragraph{The estimand is a trend difference.}
For text feature $f$, year $t$ and rater
$r \in \{H, M\}$ (human, machine), let
$\beta_{r,f}(t)$ be the coefficient of the within-year standardised feature on the
within-year standardised score, estimated with the other features and abstract length
mutually controlled. An observed trajectory $\beta_{H,f}(t)$ moves for two reasons,
\begin{equation}
\underbrace{\Delta \beta_{H,f}}_{\text{observed}} \;=\;
\underbrace{C_f}_{\text{composition}} \;+\;
\underbrace{P_f}_{\text{preference}} ,
\end{equation}
and the frozen arm gives $\Delta \beta_{M,f} = C_f$ alone, because the machine's scoring
rule is literally the same function in every year. The estimand is therefore the trend
difference
\begin{equation}
P_f \;=\; \Delta \beta_{H,f} - \Delta \beta_{M,f},
\end{equation}
which we implement as the three-way interaction
$\text{feature} \times \text{year} \times \mathbb{1}[\text{human}]$ on the stacked human and
machine observations of the same papers, the multi-period difference-in-differences of
\citet{callaway2021did} with a rater rather than a treatment group as the second
difference.

\paragraph{What must be true.}
Three conditions carry the design. (i)~\emph{The rater is frozen.} All machine reviews come
from one model family, one prompt, one short window; this is a property of how the corpus
was built, stated by its authors \citep{demetrio2025genreview}, not an assumption we impose.
It is the reason the design works here and would not work against a live API endpoint, whose
silent updates are themselves a documented source of measurement change
\citep{yang2026judge,li2026drifted}. (ii)~\emph{Both arms see the same papers.} We use only
submissions with both a human and a machine score, so composition is held identical across
arms by construction. (iii)~\emph{Composition acts on both arms in the same direction.} We
do not require equal magnitudes; we require that a change in the mix of texts not move the
two raters' coefficients in opposite directions. This is the one assumption that is not
guaranteed by construction, and Section~\ref{sec:placebo} tests it with placebos.

\paragraph{What is not assumed.}
We do not assume the machine agrees with humans, is accurate, or should be imitated. A
frozen rater that is systematically wrong still identifies $P_f$, provided it is wrong in
the same way every year. In the vocabulary of measurement invariance
\citep{mellenbergh1989item,holland1993dif,vandenberg2000review}, the frozen arm is the anchor
assumed invariant, and the quantity of interest is a slope difference across groups, which is
non-uniform differential item functioning \citep{swaminathan1990detecting}.

\paragraph{What the design cannot do.}
It cannot say whether the human shift is an improvement. It measures that the humans moved
and the fixed instrument did not. Normative content enters only in
Section~\ref{sec:discussion}, and only as an interpretation marked as such.
Table~\ref{tab:design} sets the design against the correlational trajectory previously
reported on this venue.

\begin{table}[t]
\centering
\small
\begin{tabular}{p{2.9cm}p{5.6cm}p{6.6cm}}
\toprule
 & Correlational trajectory \citep{jung2025acceptance} & Frozen-rater design (this paper) \\
\midrule
Outcome & acceptance & human review score, standardised within year \\
Text layer & full text as currently hosted (post-revision) & abstract as submitted (rebuilt from version histories) \\
Estimator & univariate correlations & four cues and abstract length mutually controlled; paper-clustered SE \\
Time resolution & two bins, before and after the LLM release & eight yearly points and an early-to-late contrast \\
Comparator & none & a rater frozen by construction, scored on the same papers \\
Estimand & association between readability and acceptance & human-minus-frozen trend difference, i.e.\ preference drift net of composition drift \\
Error control & none reported & BH-FDR $q<0.10$ and a 90\% interval, both required; forty placebo wordlists; year permutation \\
\bottomrule
\end{tabular}

\caption{\textbf{Two ways to read a moving association.} The correlational design reports
that the association changed; the frozen-rater design attributes the change to the
evaluator or to the corpus.}
\label{tab:design}
\end{table}

\subsection{Estimation and the double gate}
\label{sec:estimation}

Within each year and arm we regress the standardised score on the four standardised cues,
cluster by paper, and read off $\beta_{r,f}(t)$. We contrast an early window (2018--2020)
with a late one (2023--2025) on the same specification, and we run the pre-period check
recommended for difference-in-differences designs \citep{roth2023trending} on the
2018--2021 window (Section~\ref{sec:placebo}). The trend statistic is the three-way
interaction of Section~\ref{sec:ident}. We report effect sizes as the change in score, in
score standard deviations, from moving a cue from its median to its 90th percentile, with
bootstrap intervals at $B=1{,}500$ \citep{efron1979bootstrap}.

\paragraph{Every claim passes a double gate.}
We require every claim to pass \emph{both} BH-FDR at $q<0.10$ \citep{benjamini1995controlling}
and a 90\% interval excluding zero. Of $23$ pre-specified tests, $12$ pass. We report the
other $11$ as not passing, mark them in every table and plot, and list all $23$ in
Appendix~\ref{app:inference}.

\paragraph{Placebos supply the null distribution.}
Because the estimand is a contrast between two arms on the same units, its natural
null distribution comes from re-running the identical specification on cues that carry no
signal, in the spirit of placebo inference for comparative designs
\citep{abadie2010synthetic}: forty random mid-frequency wordlists treated as if each were
the cue of interest, and thirty permutations of the year labels.

\paragraph{One specification trap, since it produced a false positive for us.}
Including a feature-by-centred-year interaction together with year fixed effects makes the
design matrix rank-deficient. The HC3 covariance estimator \citep{mackinnon1985hc3} does
not fail loudly in that situation; it returns standard errors of $0.0000$ and
correspondingly spectacular significance. Our robustness code now asserts full rank before
reporting.

\section{Results}
\label{sec:results}

With the panel, the cues and the decision rule in place, we turn to the estimates.

\subsection{The human weight on non-domain lexical complexity fell to zero; the frozen rater did not move}
Table~\ref{tab:yearly} gives the year-by-year coefficients on the same papers. The human
coefficient on non-domain lexical complexity runs $+0.142$, $+0.081$, $+0.018$, $+0.060$,
$+0.080$, $+0.015$, $+0.024$, $-0.015$ from 2018 to 2025; the frozen rater runs $+0.080$,
$+0.077$, $+0.101$, $+0.060$, $+0.113$, $+0.071$, $+0.063$, $+0.082$. Both series are noisy
year to year and their per-year intervals overlap; the claim rests on the trend difference,
tested below, not on any single year. Figure~\ref{fig:main} shows the shape: the two arms
track each other until 2022 and separate afterward, while the corpus mean of the cue
(Figure~\ref{fig:drift}) climbs most steeply in exactly the years the human coefficient sits
at zero.

\begin{table}[t]
\centering
\scriptsize
\setlength{\tabcolsep}{4pt}
\begin{tabular}{lcccccc}
\toprule
 & \multicolumn{2}{c}{Non-domain lexical complexity} & \multicolumn{2}{c}{Domain jargon} & \multicolumn{2}{c}{Sentence-length SD} \\
\cmidrule(lr){2-3}\cmidrule(lr){4-5}\cmidrule(lr){6-7}
Year & human & frozen & human & frozen & human & frozen \\
\midrule
2018 & $+0.142$ (0.039) & $+0.080$ (0.039) & $+0.085$ (0.033) & $+0.047$ (0.034) & $+0.015$ (0.038) & $+0.000$ (0.038) \\
2019 & $+0.081$ (0.030) & $+0.077$ (0.030) & $+0.083$ (0.027) & $+0.001$ (0.027) & $+0.043$ (0.025) & $-0.022$ (0.026) \\
2020 & $+0.018$ (0.024) & $+0.101$ (0.024) & $+0.017$ (0.022) & $+0.001$ (0.022) & $+0.033$ (0.023) & $-0.015$ (0.024) \\
2021 & $+0.060$ (0.022) & $+0.060$ (0.023) & $+0.044$ (0.020) & $+0.037$ (0.020) & $+0.052$ (0.021) & $-0.014$ (0.021) \\
2022 & $+0.080$ (0.022) & $+0.113$ (0.023) & $+0.059$ (0.020) & $+0.012$ (0.021) & $+0.033$ (0.021) & $-0.033$ (0.022) \\
2023 & $+0.015$ (0.018) & $+0.071$ (0.019) & $+0.050$ (0.017) & $+0.003$ (0.017) & $+0.043$ (0.016) & $+0.029$ (0.016) \\
2024 & $+0.024$ (0.012) & $+0.063$ (0.014) & $+0.002$ (0.012) & $+0.020$ (0.014) & $+0.050$ (0.012) & $-0.004$ (0.013) \\
2025 & $-0.015$ (0.009) & $+0.082$ (0.011) & $-0.013$ (0.010) & $+0.018$ (0.011) & $+0.030$ (0.010) & $-0.014$ (0.012) \\
\bottomrule
\end{tabular}

\caption{\textbf{Per-year coefficients on the review score, by cue and rater.} Entries are
$\beta$ (SE) of the within-year standardised cue on the within-year standardised score, with
the other cues and abstract length controlled and standard errors clustered by paper.}
\label{tab:yearly}
\end{table}

\subsection{The decline is a lexical-complexity phenomenon, not a general one}
\begin{figure}[t]
\centering
\includegraphics[width=\textwidth]{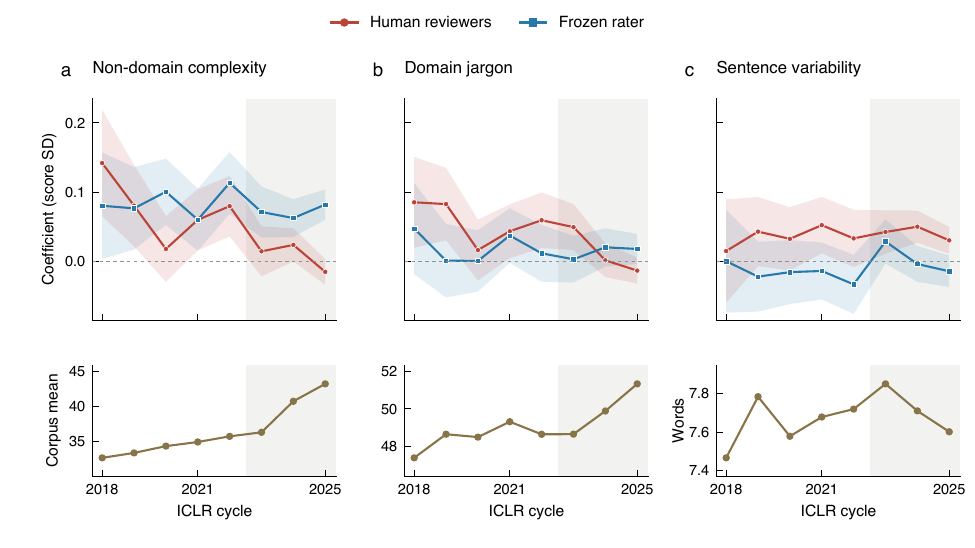}
\caption{\textbf{Cue use and corpus composition.} Top: yearly cue-score coefficients on a
shared vertical scale, with 95\% confidence bands. Bottom: corresponding corpus means in
submitted abstracts. Columns show non-domain lexical complexity, domain jargon, and
sentence-length variability. Shading denotes the 2023--2025 ICLR cycles.}
\label{fig:yearly}
\end{figure}

Figure~\ref{fig:yearly} extends the comparison to the other two cues, each over its own
corpus strip. Domain jargon follows non-domain complexity down in the human arm ($+0.085
\to -0.013$) while the frozen arm stays near a low constant. Sentence-length variability
reverses the ordering: humans hold a visible premium in every year and the frozen rater sits
near zero throughout, so the human decline is confined to the two lexical cues.

\subsection{Pooling years makes the asymmetry unambiguous}
\begin{figure}[t]
\centering
\includegraphics[width=0.92\textwidth]{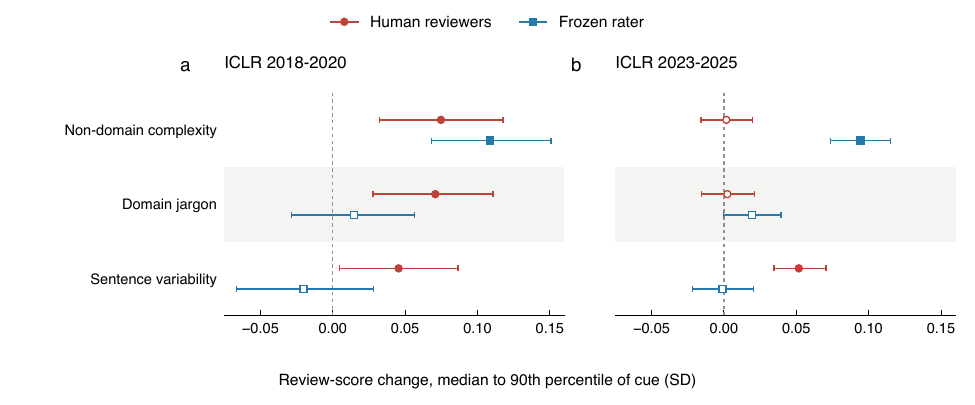}
\caption{\textbf{Effect sizes by window.} Change in review score, in score standard deviations,
from moving each cue from its median to its 90th percentile, for the early and late windows and
both raters. Whiskers are bootstrap 95\% intervals; filled markers indicate intervals excluding
zero. Non-domain complexity and sentence variability abbreviate non-domain lexical complexity
and sentence-length variability.}
\label{fig:earlylate}
\end{figure}

Figure~\ref{fig:earlylate} and Table~\ref{tab:earlylate} give the compact form of the result.
Moving non-domain lexical complexity from its median to its 90th percentile was worth
$+0.075$ score SD to human reviewers in 2018--2020 and $+0.002$ in 2023--2025; for the frozen
rater the same manipulation is worth $+0.109$ then and $+0.094$ now. Domain jargon goes
$+0.071 \to +0.002$ for humans and $+0.015 \to +0.019$ for the machine. Sentence-length
variability goes $+0.045 \to +0.052$ for humans and $-0.020 \to -0.001$ for the machine.

\begin{table}[t]
\centering
\scriptsize
\setlength{\tabcolsep}{4pt}
\begin{tabular}{llcccc c}
\toprule
 & & \multicolumn{2}{c}{2018--2020} & \multicolumn{2}{c}{2023--2025} & \\
\cmidrule(lr){3-4}\cmidrule(lr){5-6}
Cue & Rater & effect & 95\% CI & effect & 95\% CI & change \\
\midrule
Non-domain lexical complexity & human & $+0.075$ & [+0.032, +0.118] & $+0.002$ & [-0.016, +0.020] & $-0.073$ \\
 & frozen & $+0.109$ & [+0.068, +0.151] & $+0.094$ & [+0.074, +0.115] & $-0.014$ \\
Domain jargon & human & $+0.071$ & [+0.028, +0.111] & $+0.002$ & [-0.016, +0.021] & $-0.068$ \\
 & frozen & $+0.015$ & [-0.029, +0.057] & $+0.019$ & [-0.000, +0.039] & $+0.005$ \\
Sentence-length SD & human & $+0.045$ & [+0.005, +0.087] & $+0.052$ & [+0.035, +0.071] & $+0.006$ \\
 & frozen & $-0.020$ & [-0.067, +0.028] & $-0.001$ & [-0.022, +0.020] & $+0.019$ \\
\bottomrule
\end{tabular}

\caption{\textbf{Effect sizes by window.} Change in review score, in score SD, from moving a
cue from its median to its 90th percentile; bootstrap 95\% CIs at $B=1{,}500$.}
\label{tab:earlylate}
\end{table}

\paragraph{Both lexical-complexity contrasts clear the double gate; the frozen arm's do
not.}
Table~\ref{tab:inference}, plotted as Figure~\ref{fig:inference} in
Appendix~\ref{app:inference}, gives the inference. The three-way DiD is $-0.0100$ for
non-domain lexical complexity ($q=0.013$) and $-0.0098$ for domain jargon ($q=0.014$);
both clear the double gate. For sentence-length variability it is $-0.0005$ ($q=0.896$) and
does not. The period contrast is the cleanest statement of the asymmetry: human $-0.0485$
($q=0.012$) and $-0.0433$ ($q=0.018$), frozen rater $-0.0076$ ($q=0.775$) and $+0.0084$
($q=0.775$). The humans moved; the fixed instrument did not.

\begin{table}[t]
\centering
\small
\begin{tabular}{lrrlcc}
\toprule
Test & $\beta$ & SE & 90\% CI & $q_{\text{BH}}$ & gate \\
\midrule
\addlinespace\multicolumn{6}{l}{\textit{DiD three-way (feature $\times$ year $\times$ human)}} \\
\quad non-domain lexical complexity & $-0.0100$ & 0.0035 & [-0.0158, -0.0042] & $0.013$ & \checkmark \\
\quad domain jargon & $-0.0098$ & 0.0035 & [-0.0155, -0.0040] & $0.014$ & \checkmark \\
\quad sentence-length SD & $-0.0005$ & 0.0037 & [-0.0066, +0.0056] & $0.896$ & -- \\
\addlinespace\multicolumn{6}{l}{\textit{2023--2025 cross-section}} \\
\quad human, sentence-length SD & $+0.0396$ & 0.0069 & [+0.0282, +0.0509] & $<0.001$ & \checkmark \\
\quad frozen, non-domain lexical complexity & $+0.0715$ & 0.0076 & [+0.0590, +0.0841] & $<0.001$ & \checkmark \\
\addlinespace\multicolumn{6}{l}{\textit{Early$\to$late contrast}} \\
\quad human, non-domain lexical complexity & $-0.0485$ & 0.0166 & [-0.0759, -0.0211] & $0.012$ & \checkmark \\
\quad human, domain jargon & $-0.0433$ & 0.0163 & [-0.0702, -0.0165] & $0.018$ & \checkmark \\
\quad frozen, non-domain lexical complexity & $-0.0076$ & 0.0163 & [-0.0345, +0.0192] & $0.775$ & -- \\
\quad frozen, domain jargon & $+0.0084$ & 0.0170 & [-0.0195, +0.0363] & $0.775$ & -- \\
\bottomrule
\end{tabular}

\caption{\textbf{Main inference.} A claim is made only when BH-FDR $q<0.10$ and the 90\%
interval excludes zero (\checkmark). Rows without the mark fail the gate and are reported as
failures; all $23$ pre-specified tests are in Table~\ref{tab:inference_full}.}
\label{tab:inference}
\end{table}

\subsection{By 2023--2025 the two raters price different things}
Humans and the frozen rater agree on totals and disagree on reasons. In the late window
humans put $+0.0396$ ($q<0.001$) on sentence-length variability, a cue the frozen rater does
not register, and the frozen rater puts $+0.0715$ ($q<0.001$) on non-domain lexical
complexity, a cue humans no longer pay for. The overall score correlation between the arms
is $0.245$, close to the human--human agreement reported in the meta-analytic literature on
journal peer review (ICC around $0.34$, \citealp{bornmann2010reliability}) and to the
disagreement documented in the venue's own consistency experiments
\citep{cortes2021inconsistency,beygelzimer2023consistency}. The two raters agree about as
well as two humans do while weighting the observable text quite differently: agreement at
the level of totals is not agreement about reasons.

\subsection{Forty random wordlists do not reproduce the decline}
\label{sec:placebo}

\begin{figure}[t]
\centering
\includegraphics[width=0.62\textwidth]{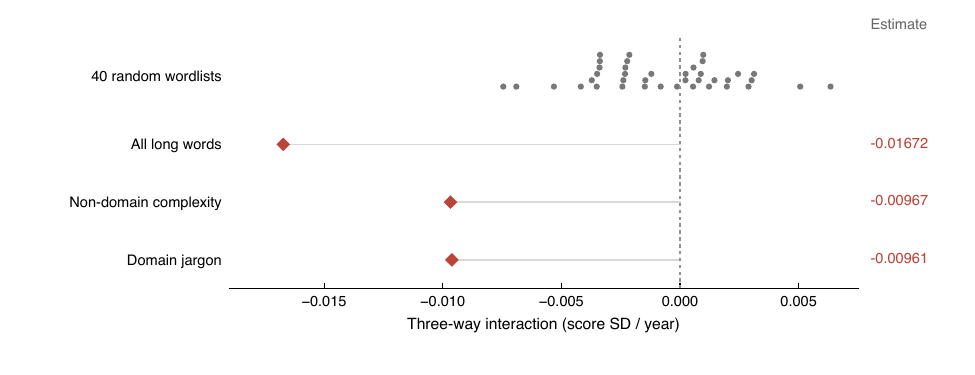}
\caption{\textbf{Observed interactions relative to random wordlists.} Each grey dot is one of
forty placebo interactions; vertical stacking prevents overlap. Red diamonds show three real
lexical-cue estimates in the same specification, with separate rows for the nearly coincident
non-domain and domain estimates. Grey stems join estimates to zero; they are not confidence
intervals. The all-long-words measure is not sentence-length variability.}
\label{fig:placebo}
\end{figure}

\paragraph{The decline is cue-specific.}
The threat is that every coefficient drifts toward zero for reasons unrelated to the cue:
more reviews per paper, a changing reviewer pool, score compression. We built $40$ random
mid-frequency wordlists, treated each as if it were the cue of interest, and ran it through
the identical three-way specification (Table~\ref{tab:robust}). They centre on $-0.0005$
with SD $0.0033$ (5th and 95th percentiles $-0.0054$ and $+0.0051$). The real cues sit at
$-0.0097$ (non-domain lexical complexity), $-0.0096$ (domain jargon) and $-0.0167$ (all
long words), outside the whole placebo distribution, empirical $p = 0.000$ in each case
(Figure~\ref{fig:placebo}). Generic shrinkage would have moved the placebos too. This is also
the test of condition~(iii) in Section~\ref{sec:ident}: a composition change that pushed
the two arms in opposite directions for reasons unrelated to the cue would have to do so
for the real cues and not for forty random ones.

\begin{table}[t]
\centering
\scriptsize
\setlength{\tabcolsep}{4pt}
\begin{tabular}{llrrl}
\toprule
Check & Statistic & Estimate & SD / SE & Interval or $p$ \\
\midrule
Random-wordlist placebos ($n=40$) & three-way DiD & $-0.0005$ & 0.0033 & [-0.0054, +0.0051] \\
\quad real cue: non-domain lexical complexity & three-way DiD & $-0.0097$ & -- & empirical $p=0.000$ \\
\quad real cue: domain jargon & three-way DiD & $-0.0096$ & -- & empirical $p=0.000$ \\
\quad real cue: all long words & three-way DiD & $-0.0167$ & -- & empirical $p=0.000$ \\
Year-label permutations ($n=30$) & three-way DiD & $-0.0004$ & 0.0040 & share $\geq |\text{real}|$: 0.000 \\
$R^2$ trend, human arm & slope per year & $-0.00183$ & 0.0007 & [-0.00355, -0.00011] \\
$R^2$ trend, frozen arm & slope per year & $+0.00017$ & 0.00054 & [-0.00114, +0.00148] \\
$R^2$ trend, difference & slope per year & $-0.002$ & 0.0009 & [-0.0042, +0.0002] \\
\bottomrule
\end{tabular}

\caption{\textbf{Robustness of the three-way interaction.} Placebo wordlists and year
permutations are run through the identical specification; $R^2$ trends are two-parameter
fits to eight yearly points and are reported as diagnostics, not evidence.}
\label{tab:robust}
\end{table}

\paragraph{Shuffling the years destroys it.}
Shuffling year labels $30$ times gives interactions centred at $-0.0004$ (SD $0.0040$);
none reaches the magnitude of the real estimate.

\paragraph{The effect sits after the cost shock, not before it.}
Restricting to 2018--2021 gives $-0.0129$ ($p=0.359$); 2022--2025 gives $-0.0353$
($p<0.001$). The effect is concentrated where the cost shock is, which is also visible in
Figure~\ref{fig:main}.

\subsection{Explained variance is reported, not relied on}
Table~\ref{tab:r2} and Figure~\ref{fig:r2} in Appendix~\ref{app:extra} report per-year
$R^2$ for both arms. We initially read the
endpoints ($0.0197 \to 0.0019$ for humans against a flat machine arm) as evidence against
generic shrinkage. That was reading a trend off two points. Fitting the trend on eight
observations gives $-0.00183$/yr for humans, 95\% CI $[-0.00355,-0.00011]$, and
$+0.00017$/yr for the frozen arm, CI $[-0.00114,+0.00148]$. The human interval just excludes
zero, but a two-parameter fit to eight points is not something we rest a claim on; we let
the placebo battery, which has power, carry the argument.

\subsection{Seven findings we withdrew}
\label{sec:retractions}
Four findings with attractive first-run $p$-values did not survive adversarial re-testing,
and three more fell to a control we should have applied first;
Table~\ref{tab:retractions} in Appendix~\ref{app:retractions} lists all seven with the
diagnostic that overturned each. The common cause was interpreting before testing, and the
failure modes recur in text-and-outcomes work: a numeral regex that matched model names
rather than results; a hype list and an LLM-marker list that were mostly ordinary
vocabulary; an uncontrolled complexity penalty that went to zero, not negative, under
controls; a rotation of the human evaluation function estimated on a single year; and a
presentation-versus-soundness dissociation that was a team-size confound. The procedural
lessons are now rules in our pipeline: any regex or wordlist measure must first emit its
twenty most frequent hits in context; any vector estimated on a small sample must be
bootstrapped; any obvious structural covariate must be controlled before interpretation.

\section{Discussion, limitations, and conclusion}
\label{sec:discussion}

The estimates answer the three questions in order: the human weight on lexical complexity
moved (RQ1), the movement is confined to the cues whose cost collapsed (RQ2), and the frozen
rater still pays for exactly those cues (RQ3). We now say what that does and does not
license.

\paragraph{Reviewers discounted a cue whose cost collapsed.}
Human reviewers reduced the weight they place on a cue whose production cost collapsed.
Models of evaluation under manipulable signals prescribe exactly that comparative static:
underweight what the sender can cheaply produce, and underweight it more as the ability to
produce it spreads. We say the observed movement is \emph{consistent with} that
prescription. We cannot show reviewers arrived at the optimum, nor that they were
deliberate; a taste change with the same sign would look identical in our data. The timing
is informative: the human coefficient sits at zero from ICLR 2023 onward, a cycle whose
papers were written before the release of ChatGPT but reviewed after it, while the corpus
mean of the cue jumps only from ICLR 2024, the first cycle written after it. The reviewers
moved before the papers did.

\paragraph{The frozen rater is the cautionary half.}
The same property that makes the machine a valid instrument makes it a poor judge. It has
preserved its earlier reward schedule into 2025 and still pays $+0.0715$ for a cue that
human reviewers have written down to zero, while remaining blind to the structural feature
humans now price most. Its total-score agreement with humans, $0.245$, is close to
human--human agreement, so the misalignment is invisible from agreement statistics alone:
right answer, wrong reasons. Deployments that freeze a prompt for reproducibility should
expect this drift to accumulate silently, and should measure their rater's feature weights
rather than only its correlation with humans. Acting on a published coefficient is
self-defeating for the same reason the coefficient moved: the cue loses value as it is
optimised \citep{perdomo2020performative}.

\paragraph{Calibrating a judge to historical preferences bakes in the debased cue.}
Any LLM judge that is prompted, fine-tuned or reward-modelled to reproduce human review
scores learns the human reward schedule of its calibration period. Our estimates imply a
testable ordering: the earlier the calibration window, the larger the weight such a judge
places on non-domain lexical complexity relative to contemporary reviewers, and the larger
the score gain available from rewriting alone. The frozen rater studied here is the limiting
case of a judge calibrated once and never updated; judges refreshed on recent reviews should
show a smaller gap, and the gap should be measurable with the same design.

\paragraph{Limitations.}
Abstracts, not full text, for the reasons given: no unbiased full-text route exists at this
scale. One venue. Eight years, and the interesting part is the last three. Our cues are
lexical and structural proxies for something vaguer. The machine rater is one model family
under one prompt, so ``the frozen rater'' is that configuration and not LLMs in general.
Condition~(iii) of Section~\ref{sec:ident} is tested, not guaranteed. Human ratings
themselves are noisy, with reported inter-reviewer ICC around $0.34$. Two version caveats
remain after the rebuild: the 2024 abstracts are post-discussion rather than submitted
text, because no earlier public snapshot exists, and the machine rater scored current PDFs,
so its coefficient \emph{levels} carry a version effect even though its year-to-year
flatness does not. Finally, the divergence says nothing about whether any paper was judged
correctly.

\paragraph{The design applies wherever a fixed-prompt rater exists.}
Any setting with a fixed-prompt automated rater applied retrospectively across a period in
which human standards may have moved qualifies: grading, content moderation, hiring
screens, clinical scoring. The requirement is strict and easy to violate. The rater must not
be updated, re-prompted, or silently re-routed mid-study; a live API endpoint does not
satisfy it.

\paragraph{Conclusion.}
A rater that never changes is useless as a judge and valuable as an instrument. Holding one
still, we can see that human reviewers stopped paying for lexical complexity while the
machine kept paying. The methodological point is the one to carry: stable automated raters
can diagnose human preference change, and should not thereby be mistaken for the standard.
The question is no longer whether reviewers reward lexical complexity, but whose standard
moved, and by how much.

\section{Reproducibility}
\label{sec:repro}

\paragraph{Data availability.}
All inputs are public. Human reviews, decisions and abstract version histories come from
OpenReview (\url{https://openreview.net}; API v1 for 2018--2023, API v2 thereafter). The
machine reviews are the Gen-Review corpus \citep{demetrio2025genreview}, released at
\url{https://anonymous.4open.science/r/gen_review/}. Pre-review abstract snapshots for
ICLR 2024 and 2025 come from \url{https://github.com/berenslab/iclr-dataset}. Syllable
counts use the CMU pronouncing dictionary. Because the venue's anonymous query endpoint
began returning HTTP 403 during this work, we also preserved a dataset snapshot and report
its SHA-256 so that our inputs remain checkable.

\paragraph{Code availability.}
The full pipeline is released at \url{https://github.com/Biajin-PKU/frozen-rater-drift}:
corpus assembly and the abstract-version rebuild, feature construction with both wordlists
and their hit distributions, estimation, the $40$ placebo draws and $30$ permutation draws,
the diagnostic scripts behind each of the seven withdrawn findings, and the scripts that
render every table and figure. Every number in this paper, every table and every rendered
figure is drawn from a single machine-readable records file in that repository, and a
checker verifies that no numeral in the manuscript is absent from it.

\paragraph{Rater configuration.}
Model family, generation window, prompt, and decoding settings for the frozen arm are those
of the Gen-Review release; verifying that the rater was fixed requires only those fields,
and we reproduce them with the released artefacts.

\paragraph{Author contributions (CRediT).}
\textbf{Jiabin Zheng:} Conceptualization, Methodology, Software, Validation, Formal analysis, Investigation, Data curation, Writing -- original draft, Writing -- review \& editing, Visualization.
The author has read and agreed to the submitted version of the manuscript.

\bibliographystyle{plainnat}
\bibliography{refs}

\appendix
\section{All pre-specified tests}
\label{app:inference}

\begin{table}[H]
\centering
\small
\setlength{\tabcolsep}{4pt}
\begin{tabular}{lrrlcc}
\toprule
Test & $\beta$ & SE & 90\% CI & $q_{\text{BH}}$ & gate \\
\midrule
\addlinespace\multicolumn{6}{l}{\textit{DiD three-way (feature $\times$ year $\times$ human)}} \\
\quad non-domain lexical complexity & $-0.0100$ & 0.0035 & [-0.0158, -0.0042] & $0.013$ & \checkmark \\
\quad domain jargon & $-0.0098$ & 0.0035 & [-0.0155, -0.0040] & $0.014$ & \checkmark \\
\quad sentence-length SD & $-0.0005$ & 0.0037 & [-0.0066, +0.0056] & $0.896$ & -- \\
\addlinespace\multicolumn{6}{l}{\textit{2023--2025 cross-section}} \\
\quad human, non-domain lexical complexity & $+0.0014$ & 0.0069 & [-0.0099, +0.0127] & $0.896$ & -- \\
\quad human, domain jargon & $+0.0019$ & 0.0070 & [-0.0096, +0.0133] & $0.896$ & -- \\
\quad human, sentence-length SD & $+0.0396$ & 0.0069 & [+0.0282, +0.0509] & $<0.001$ & \checkmark \\
\quad frozen, non-domain lexical complexity & $+0.0715$ & 0.0076 & [+0.0590, +0.0841] & $<0.001$ & \checkmark \\
\quad frozen, domain jargon & $+0.0164$ & 0.0077 & [+0.0037, +0.0291] & $0.065$ & \checkmark \\
\quad frozen, sentence-length SD & $-0.0014$ & 0.0082 & [-0.0149, +0.0121] & $0.896$ & -- \\
\addlinespace\multicolumn{6}{l}{\textit{Early$\to$late contrast}} \\
\quad human, non-domain lexical complexity & $-0.0485$ & 0.0166 & [-0.0759, -0.0211] & $0.012$ & \checkmark \\
\quad human, domain jargon & $-0.0433$ & 0.0163 & [-0.0702, -0.0165] & $0.018$ & \checkmark \\
\quad human, sentence-length SD & $+0.0145$ & 0.0161 & [-0.0120, +0.0409] & $0.607$ & -- \\
\quad frozen, non-domain lexical complexity & $-0.0076$ & 0.0163 & [-0.0345, +0.0192] & $0.775$ & -- \\
\quad frozen, domain jargon & $+0.0084$ & 0.0170 & [-0.0195, +0.0363] & $0.775$ & -- \\
\quad frozen, sentence-length SD & $+0.0131$ & 0.0180 & [-0.0165, +0.0426] & $0.688$ & -- \\
\addlinespace\multicolumn{6}{l}{\textit{Subscores 2024--2025}} \\
\quad presentation, non-domain lexical complexity & $+0.0192$ & 0.0076 & [+0.0067, +0.0317] & $0.024$ & \checkmark \\
\quad presentation, domain jargon & $-0.0243$ & 0.0076 & [-0.0368, -0.0118] & $0.005$ & \checkmark \\
\quad soundness, non-domain lexical complexity & $-0.0247$ & 0.0076 & [-0.0372, -0.0123] & $0.005$ & \checkmark \\
\quad soundness, domain jargon & $-0.0053$ & 0.0075 & [-0.0177, +0.0070] & $0.688$ & -- \\
\addlinespace\multicolumn{6}{l}{\textit{Subscores 2024--2025, author count controlled}} \\
\quad presentation, non-domain lexical complexity & $-0.0036$ & 0.0075 & [-0.0160, +0.0088] & $0.775$ & -- \\
\quad presentation, domain jargon & $-0.0274$ & 0.0075 & [-0.0397, -0.0151] & $0.001$ & \checkmark \\
\quad soundness, non-domain lexical complexity & $-0.0416$ & 0.0076 & [-0.0541, -0.0292] & $<0.001$ & \checkmark \\
\quad soundness, domain jargon & $-0.0076$ & 0.0075 & [-0.0199, +0.0047] & $0.546$ & -- \\
\bottomrule
\end{tabular}

\caption{\textbf{The $23$ pre-specified tests.} \checkmark\ marks tests that pass both
BH-FDR $q<0.10$ and a 90\% interval excluding zero. The subscore families test whether the
presentation and soundness subscores available from 2024 dissociate on the two lexical cues;
they do not once author count is controlled (Table~\ref{tab:retractions}, row 7).}
\label{tab:inference_full}
\end{table}

\begin{figure}[H]
\centering
\includegraphics[width=\textwidth]{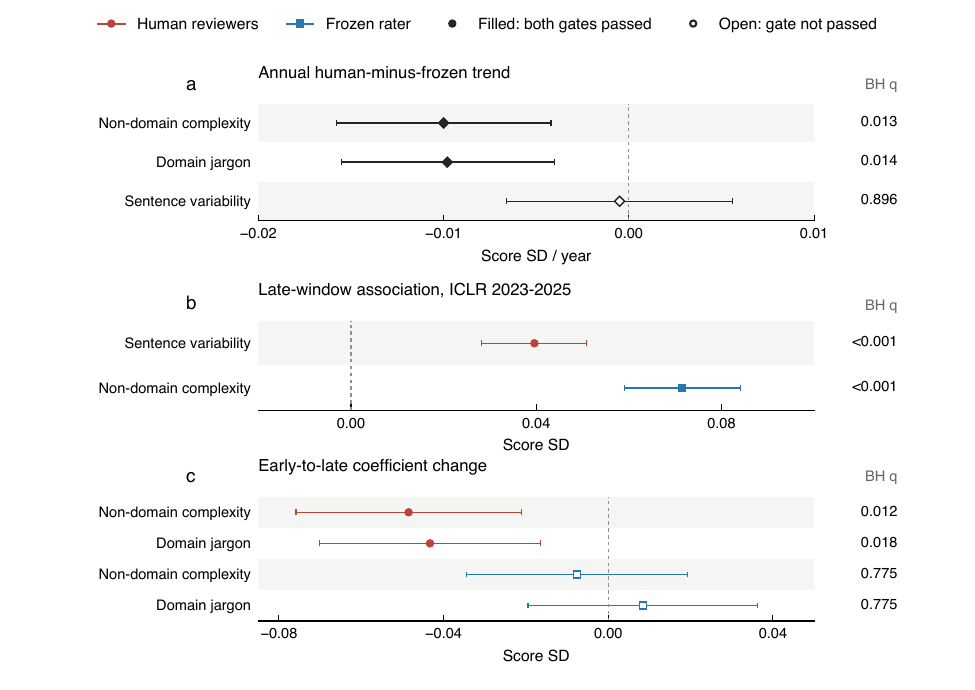}
\caption{\textbf{Headline inference, separated by estimand.} The nine headline rows of
Table~\ref{tab:inference}, with 90\% intervals and BH-adjusted $q$ values. Annual trend
differences, late-window associations, and early-to-late coefficient changes use separate scales
and units. Colour and shape distinguish raters; filled markers pass both BH-FDR $q<0.10$ and
interval exclusion, and open markers do not.}
\label{fig:inference}
\end{figure}

\section{Withdrawn findings}
\label{app:retractions}

\begin{table}[H]
\centering
\scriptsize
\begin{tabular}{p{0.5cm}p{4.2cm}p{4.6cm}p{5.4cm}}
\toprule
 & Claim on first run & Diagnostic that overturned it & Outcome \\
\midrule
1 & numerals in the abstract help ($+2.22$pp, $p=5.1\times10^{-7}$) & the regex matched model names (GPT-\textbf{4}o, Gemini-\textbf{1.5}) & result numerals $\beta=+0.016$, $p=0.17$; model-name numerals $+0.055$, $p=2.2\times10^{-6}$ \\
2 & promotional language is penalised ($-0.81$pp) & $71.8\%$ of the hype list was standard vocabulary; only $2.2\%$ of \emph{significant} was statistical & genuine promotional words $p=0.27$ \\
3 & LLM marker words are penalised & $75.4\%$ of the list was ordinary vocabulary & distinctive subset grew $+625\%$ with $\beta=-0.000$, $p=0.98$ \\
4 & complexity became a penalty & uncontrolled $-0.030$ versus controlled & $+0.006$, $p=0.63$: the weight went to zero, not negative \\
5 & the human evaluation function rotated negative ($r=-0.335$) & bootstrap interval $[-0.746,+0.301]$ & pooled years $+0.364$ \\
6 & sentence-length variability is strengthening & the human-specific level holds; the trend does not & $+0.0145$, interval covering zero \\
7 & presentation and soundness subscores dissociate & log author count predicts the presentation subscore ($+0.170$, $p=2.6\times10^{-110}$) and correlates with the cue ($+0.124$) & controlled effect $-0.0036$ from $+0.0192$ \\
\bottomrule
\end{tabular}

\caption{\textbf{Seven findings withdrawn after adversarial re-testing.} Each row gives the
first-run claim, the diagnostic that overturned it, and the outcome. The diagnostic scripts
are released.}
\label{tab:retractions}
\end{table}

\section{Abstract versions}
\label{app:versions}

\begin{table}[H]
\centering
\small
\begin{tabular}{lp{5.2cm}rp{4.2cm}}
\toprule
Years & Source of the abstract text used & Submissions & Version caveat \\
\midrule
2018--2023 & first archived version from the venue's version history & $13{,}462$ & submitted text \\
2024 & earliest public snapshot & $7{,}304$ & after discussion, before camera-ready \\
2025 & public pre-review scrape & $10{,}079$ & submitted text \\
all & current text retained where no earlier version exists (flagged) & $1{,}779$ & current \\
\midrule
\multicolumn{4}{p{14.5cm}}{Audit on $1{,}800$ stratified submissions from 2018--2023: among abstracts that changed between first and last version, the released text matched the last version in every case; the share that changed rose from $12\%$ (2018) to $72\%$ (2021); median similarity of changed abstracts to the submitted text $0.73$. Early-to-late human contrast on non-domain lexical complexity: $-0.0433$ on current text, $-0.0485$ on submitted text.} \\
\bottomrule
\end{tabular}

\caption{\textbf{Which text the cues were computed on.} The rebuilt abstract layer and the
version audit that motivated it.}
\label{tab:versions}
\end{table}

\section{Explained variance}
\label{app:extra}

\begin{table}[H]
\centering
\small
\begin{tabular}{lrrrr}
\toprule
Year & $R^2$ human & $n$ & $R^2$ frozen & $n$ \\
\midrule
2018 & 0.0197 & $922$ & 0.0059 & $925$ \\
2019 & 0.0123 & $1{,}419$ & 0.0069 & $1{,}390$ \\
2020 & 0.0013 & $2{,}213$ & 0.0084 & $2{,}175$ \\
2021 & 0.0063 & $2{,}594$ & 0.0043 & $2{,}526$ \\
2022 & 0.0083 & $2{,}618$ & 0.0135 & $2{,}522$ \\
2023 & 0.0064 & $3{,}797$ & 0.0047 & $3{,}662$ \\
2024 & 0.0030 & $7{,}262$ & 0.0043 & $5{,}607$ \\
2025 & 0.0019 & $11{,}517$ & 0.0101 & $8{,}375$ \\
\bottomrule
\end{tabular}

\caption{\textbf{Explained variance of the four-cue model by year and rater.}}
\label{tab:r2}
\end{table}

\begin{figure}[H]
\centering
\includegraphics[width=0.52\textwidth]{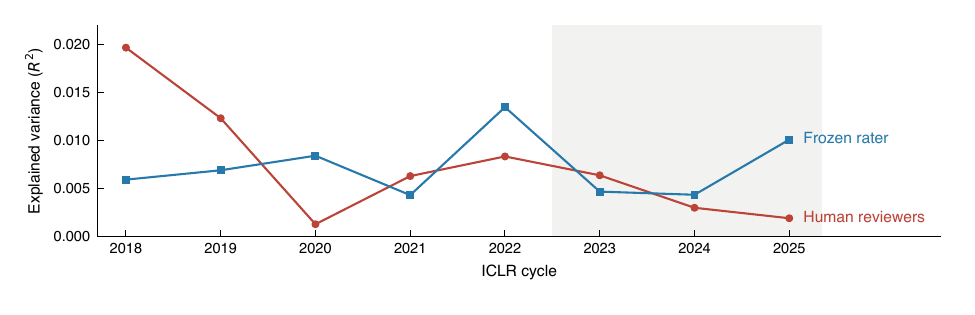}
\caption{\textbf{Explained variance by ICLR cycle.} Observed yearly $R^2$ for the four-cue
model; lines join the eight observations and are not fitted trends. Shading marks the 2023--2025
cycles. This is a descriptive diagnostic, not an identification test.}
\label{fig:r2}
\end{figure}

\end{document}